\def\year{2022}\relax
\documentclass[letterpaper]{article} 
\usepackage{aaai22}  
\usepackage{times}  
\usepackage{helvet}  
\usepackage{courier}  
\usepackage[hyphens]{url}  
\usepackage{graphicx} 
\usepackage{natbib}  
\usepackage{caption} 
\DeclareCaptionStyle{ruled}{labelfont=normalfont,labelsep=colon,strut=off} 
\usepackage{algorithm}
\usepackage{algorithmic}

\usepackage{newfloat}
\usepackage{listings}
\floatstyle{ruled}
\newfloat{listing}{tb}{lst}{}
\floatname{listing}{Listing}

\usepackage{amsmath,amssymb,amsfonts}
\usepackage{multirow}
\usepackage{makecell}
\usepackage{hhline}
\usepackage{subcaption}

\title{Uncertainty-Aware Semi-Supervised Few Shot Segmentation}
\author{
	 Soopil Kim,  Philip Chikontwe, Sang Hyun Park
}
\affiliations{
	Department of Robotics Engineering, DGIST\\
	soopilkim, philipchicco, shpark13135@dgist.ac.kr
}

\usepackage{bibentry}

\usepackage{color}

\begin{document}

\maketitle

\begin{abstract}
Few shot segmentation (FSS) aims to learn pixel-level classification of a target object in a query image using only a few annotated support samples. This is challenging as it requires modeling appearance variations of target objects and the diverse visual cues between query and support images with limited information. To address this problem, we propose a semi-supervised FSS strategy that leverages additional prototypes from unlabeled images with uncertainty guided pseudo label refinement. To obtain reliable prototypes from unlabeled images, we meta-train a neural network to jointly predict segmentation and estimate the uncertainty of predictions.  We employ the uncertainty estimates to exclude predictions with high degrees of uncertainty for pseudo label construction to obtain additional prototypes based on the refined pseudo labels. During inference, query segmentation is predicted using prototypes from both support and unlabeled images including low-level features of the query images. Our approach is end-to-end and can easily supplement existing approaches without the requirement of additional training to employ unlabeled samples. Extensive experiments on PASCAL-$5^i$ and COCO-$20^i$ demonstrate that our model can effectively remove unreliable predictions to refine pseudo labels and significantly improve upon state-of-the-art performances.
\end{abstract}

\section{Introduction}
While deep-learning based segmentation models have shown impressive performance in various applications~\cite{chen2017rethinking,ronneberger2015u}, the need for large-scale labeled training data strongly limits scalability and performance to novel unseen classes and tasks with a different distribution from the training data. To address this, few-shot segmentation (FSS) has been proposed to train deep models in the low data setting using meta-learning to learn transferable knowledge across various tasks. FSS models perform dense pixel-level prediction of unseen images (\textit{queries}) guided by limited labeled samples (\textit{support images}). This is challenging due to the limited data samples for the unseen target objects and the large diverse appearances between support and queries, particularly if the training and testing classes present a large distributional shift.

For FSS, prototype-based models have been mainly proposed. Here, foreground/background (FG/BG) prototypes are defined using extracted feature maps from a support image and its pixel-annotation. To segment unseen query images, one predicts similarity between the query feature and the obtained prototypes. For example, \cite{wang2019panet} employs a single prototype strategy but was limited in the ability to represent different parts of the target object. Follow-up works address this by defining multiple prototypes via clustering algorithms to better model feature discrepancy and reduce semantic ambiguity~\cite{liu2020part,li2021adaptive,yang2020prototype}. Despite the progress, performance is still limited and suffers from scarce context in support data, particularly if FG/BG appearance in the support and query vary, leading to inaccurate segmentation.

Recent works have demonstrated that employing FG/BG features from unlabeled images can mitigate the aforementioned issues. For example,  PPNet~\cite{liu2020part} proposed a semi-supervised model that supplements part-aware prototypes using unlabeled images with superpixels. However, their approach does not make full use of the unlabeled image information, as they just refine support prototypes based on attention mechanism between support prototypes and superpixels from unlabeled images.
Consequently, only a few superpixels with high similarity to the support prototypes are mainly used, but they do not make good use of various FG/BG features with lower similarities during inference.

Instead, we propose a novel semi-supervised approach that leverages additional prototypes from unlabeled images via pseudo labels. Using pseudo label predictions from unlabeled data can further boost model performance, yet inaccurate predictions may equally deteriorate performance. To address this, our approach further refines initial pseudo labels by excluding unreliable predictions based on uncertainty estimation. Based on gaussian assumption on pixel predictions per class following \cite{kendall2017uncertainties,lakshminarayanan2017simple}, we integrate uncertainty estimation in FSS by training a neural network to model the mean and variance of outputs from a query and prototype feature pair.
Our intuition is that uncertainty in prototype-based FSS approaches may arise from varied observations between query and prototype feature pairs. Thus, we exclude unreliable predictions from pseudo labels of unlabeled images by only including those with high mean and low uncertainty predictions as pseudo labels. This also enables the model to learn better FG/BG features not present in the support data for improved segmentation. Notably, our approach can estimate uncertainty without degrading existing prototype-based FSS models and can be trained end-to-end without an additional learning process for unlabeled samples. During inference, we jointly employ the additional prototypes from unlabeled images with the existing support prototypes to segment an unseen query image. Our contributions are summarized as follows:
\begin{itemize}
\item We propose an uncertainty estimation method for prototype-based FSS which captures uncertainty of similarity between query feature and prototype pairs. Our method can reliably quantify uncertainty without degrading the baseline performance of existing FSS models.

\item We propose a semi-supervised FSS method that employs additional prototypes from unlabeled images using pseudo labels. Our approach is robust to the number of unlabeled samples employed despite the varied appearance between samples.
 
\item We empirically demonstrate the benefit of uncertainty-based pseudo-label refinement in the semi-supervised scenario with several ablations and report improvements over state-of-the-art on two FSS benchmarks, i.e., PASCAL-$5^i$ and COCO-$20^i$.
\end{itemize}

\section{Related Works}
\subsection{Few Shot Semantic Segmentation}
Existing few-shot segmentation (FSS) models use the meta-learning framework via task-based episodic training of support and query images. OSLSM~\cite{shaban2017one} is the first method to address FSS and predicts weights of a linear classifier from support data to discriminate the target object in a query image. Several follow-up works have proposed to segment the target object based on the similarity between query feature and class prototypes~\cite{dong2018few, wang2019panet}. Since then, various models have been proposed to define better prototypes. To better leverage different parts of target objects, PPNet~\cite{liu2020part}, ASGNet~\cite{li2021adaptive}, and RPMMs~\cite{yang2020prototype} proposed to use multiple prototypes obtained via K-means clustering, superpixel-guided clustering, and gaussian mixture models, respectively. VPI~\cite{wang2021variational} suggested using a probabilistic prototype rather than deterministic. On the other hand, others proposed different strategies based on single prototype to improve performance~\cite{wang2019panet,liu2020crnet,zhang2019pyramid,wang2020few,xie2021scale}. Notably, CANet~\cite{zhang2019canet} and PFENet~\cite{tian2020prior} argued that FSS models can predict better segmentation using low-level features from the encoder.
Departing from the meta-learning framework, ~\cite{boudiaf2021few} introduced a transductive approach to learn task-specific knowledge for each task with an impressive performance over prior methods.

As for the semi-supervised methods, PPNet also leverages unlabeled images for FSS and is closely related to our work. It divides the unlabeled image into superpixels and uses some superpixels to supplement support prototypes with a GNN. However, because only a few superpixels similar to the support prototypes are used, some unlabeled data information is discarded. Moreover, it requires a training process of the GNN to utilize unlabeled images. In this paper, we instead define additional prototypes from the pseudo label predictions of unlabeled images while avoiding any additional training as in PPNet.

\subsection{Pseudo Labels in Semi-Supervised Segmentation}
Pseudo labels are commonly used in semi-supervised learning methods, e.g., the \textit{teacher-student} network setting. 
In this scenario, a trained teacher network makes predictions (pseudo-labels) to guide student network training. However, incorrect/noisy predictions can affect student learning. To address this, consistency regularization between the teacher and student networks has been popularly used.~\cite{feng2020dmt,ke2019dual}
Meanwhile, several works suggested refining pseudo labels using estimated uncertainty. \cite{sedai2019uncertainty} quantified uncertainty as entropy of the averaged probabilities obtained by randomly applying dropout several times following the work of \cite{gal2016dropout}, and trained the model with soft labels guided by entropy. However, this method is computationally expensive with performance highly influenced by dropout sampling. Thus, \cite{li2020self} proposed to estimate uncertainty using multiple losses from several jigsaw puzzle sub-tasks. On the other hand, \cite{saporta2020esl} directly used the entropy of pixel-level probability as uncertainty for an unsupervised domain adaptation (UDA) task without resorting to prior ensemble methods. Though impressive, we believe improvements in UDA are due to the use of large-scale data, which makes entropy estimates feasible. Thus, the direct use of entropy in FSS may be error prone and challenging given a few data samples. Consequently, we employ an alternative formulation for uncertainty estimation applicable to the FSS task for pseudo-label refinement.   

\subsection{Uncertainty Estimation in Neural Network}
Although modern neural networks (NNs) show good performance on several tasks, \cite{guo2017calibration} reports that the predicted probability is often different from the actual observed likelihood, i.e., most of the expected probabilities are close to 0 or 1, and thus highly \textit{overconfident}. 
In order to quantify uncertainty of model prediction, bayesian neural networks (BNNs) have been proposed. BNN models calculate posterior probability when a prior of weights is given and the uncertainty can be quantified based on variational inference of output. Since the posterior of NNs is intractable, various approximations have been proposed~\cite{louizos2016structured,blundell2015weight}. For example, dropout-based methods are popular and frequently used in several applications~\cite{Gal2016UncertaintyID,kendall2017uncertainties,kendall2018multi}. On the other hand, non-bayesian approaches employ a gaussian distribution-based method, i.e., where the output is assumed to follow a gaussian distribution and the model estimates the mean and variance~\cite{lakshminarayanan2017simple}. Nevertheless, we argue the above approaches are difficult to correctly optimize NN parameters for the FSS task with only a small number of data samples. Thus, we consider gaussian process inspired techniques; along this line of work, gaussian process regression (GPR) can estimate the mean and variance of gaussian distribution, but requires a predefined kernel and incurs heavy computation in the order of $O(n+m)^3$ with $n$ and $m$ being the number of observations and target data, respectively. To address this, CNP~\cite{garnelo2018conditional} trained a neural network that aggregates information from given data samples and estimates mean and variance. As a result, it could reduce the computation of GPR and perform flexible tasks such as half-image completion. Inspired by CNP, we propose an uncertainty estimation module in our FSS framework. Our module estimates mean and variance of the gaussian distribution from a query feature and its nearest prototype. To the best of our knowledge, we are the first to propose an uncertainty estimation method in FSS.


\section{Methods}
\label{sec:method}
\subsection{Problem Setup}
\label{sec:formulation}
A few shot segmentation model $FSS_{\theta}$ parameterized by ${\theta}$ learns to segment an unseen target object in a query image $I_{q}$ when $K$ support image and label pairs $\begin{Bmatrix}{I^1_{s}, L^1_{s}}\end{Bmatrix}$, $\begin{Bmatrix}{I^2_{s}, L^2_{s}}\end{Bmatrix}$, ..., $\begin{Bmatrix}{I^K_{s}, L^K_{s}}\end{Bmatrix}$ are given. The model learns transferable knowledge from the training task set $T_{train}$ and is later applied on the test task set $T_{test}$ containing novel classes. $T_{train}$ and $T_{test}$ are sampled from the base task set $T_{base}$ where each element has images and pixel-wise annotations of different class, with  no overlap between the sets, i.e., $T_{train}\cap T_{test}=\emptyset$. Existing standard FSS methods address the following supervised learning problem:
\begin{equation}
	\label{eq::fss}
	L_{q} = FSS_{\theta}(\begin{Bmatrix}{I^k_{s}, L^k_{s}}\end{Bmatrix}^{K}_{k=1}, I_{q}).
\end{equation} 

In this work, we extend this setting to a semi-supervised learning problem using unlabeled images. Unlabeled images are relatively easy to obtain and the scarcity of support images can be complemented by the unlabeled images. Thus, given a set of unlabeled samples $I^{1}_{u}$, $I^{2}_{u}$, \dots $I^{M}_{u}$, the semi-supervised FSS problem can be formulated as:
\begin{equation}
	\label{eq::semi_fss}
	L_{q} = FSS_{\theta}(\begin{Bmatrix}{I^k_{s}, L^k_{s}}\end{Bmatrix}^{K}_{k=1},\begin{Bmatrix}{I^m_{u}, L^m_{u}}\end{Bmatrix}^{M}_{m=1}, I_{q}),
\end{equation}
where $M$ is the number of unlabeled images.

\begin{figure} [t]
	\begin{center}
		\includegraphics[width=0.9\linewidth] {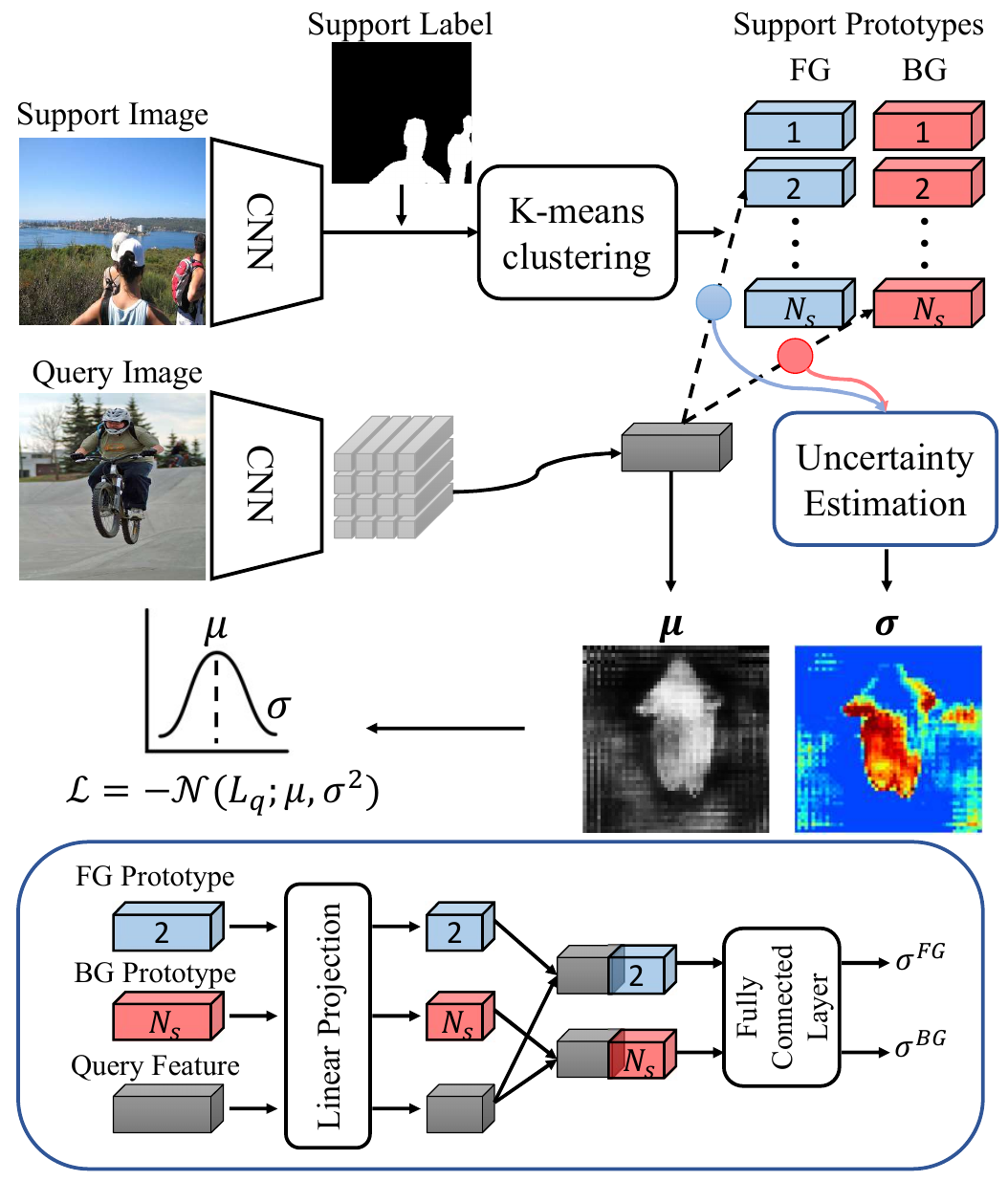}
	\end{center}
	\caption{Proposed uncertainty-aware FSS model training strategy. Blue box shows a detail of uncertainty estimation module.}
	\label{fig::train}
\end{figure}

\begin{figure*} [h]
	\begin{center}
		\includegraphics[width=0.9\linewidth] {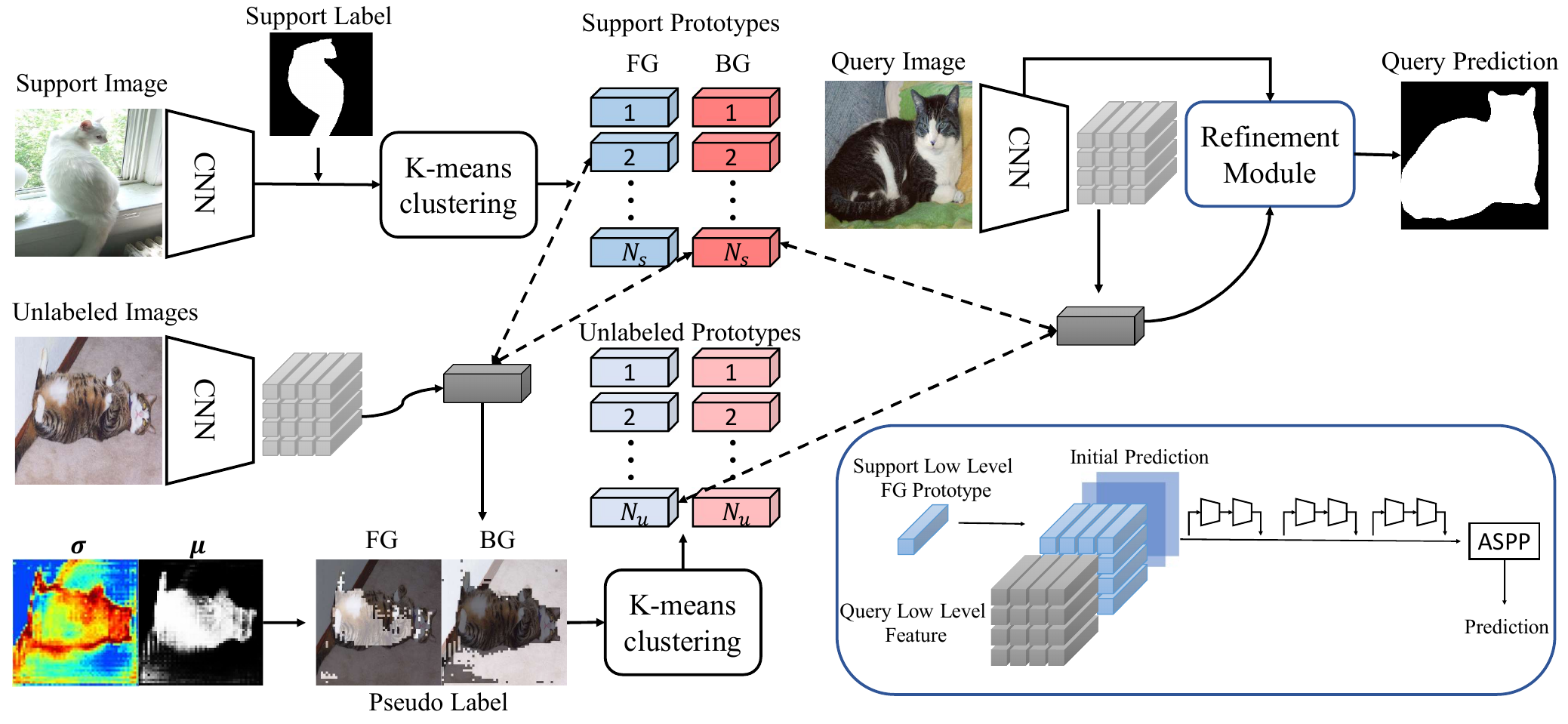}
	\end{center}
	\caption{Overview of the proposed semi-supervised FSS inference strategy. Blue box shows a detail of refinement module.}
	\label{fig::infer}
\end{figure*}

In Fig.~\ref{fig::train} and Fig.~\ref{fig::infer}, we present an overview of the training and inference pipelines of our approach. We jointly train $FSS_{\theta}$ and the uncertainty estimation module following a standard meta-training strategy without any unlabeled images. During inference, we directly employ the trained model to estimate uncertainty and refine pseudo labels for prototype generation on the unlabeled image features. Finally, we employ both the initial support and additional prototypes for segmentation of a query image. In particular, we cluster per-class support features into several clusters via K-means clustering. Here, the prototype nearest to the query feature is selected and the $\mu$ and $\sigma^2$ of the gaussian distribution are estimated using the nearest prototype-query feature pair to define pseudo labels. Following, additional prototypes are defined from the unlabeled images using the pseudo labels. For precise query segmentation, we leverage: (i) support and unlabeled prototypes, (ii) low-level features of support and query images, and (iii) the initial query prediction in a refinement module that learns cross-relations for improved segmentation. 

\subsection{Prototype-Based Few Shot Segmentation}
\label{sec:prototypefss}
This work builds upon that of PPNet~\cite{liu2020part}, a multiple prototype-based approach with a simple design. However, clear distinctions are shown by the novel modules and inference strategies introduced in our work. 

Formally, part-aware prototypes can be obtained using K-means clustering on the extracted CNN features, and later used to segment a query image based on the similarity between extracted query feature and part-aware prototypes, respectively. In particular, given a support image $I_s\in\mathbb{R}^{ W\times H\times 3}$, we obtain a feature map $f_s\in\mathbb{R}^{W'\times H'\times C}$ by feeding $I_s$ to an encoder $E_\theta$, where $C$ is channel size, $(W,H)$ and $(W',H')$ are spatial resolutions of the original image and feature maps, respectively. Here, $(W',H')$ is halved as many times as the number of max-pooling operations in $E_{\theta}$, e.g., $W’=W/8$, $H’=H/8$ with 3 max-pooling operations. At the same time, a support label $L_s$ is resized to the same size as $f_s$ and we later use this mask to separate foreground/background (FG/BG) features into $f^{fg}_s$ and $f^{bg}_s$. Using K-means clustering, features are divided into $N_s$ clusters, where $N_s$ is the number of clusters for per class-support images. Finally, the mean vectors of the features belonging to each cluster are defined as part-aware prototypes $\begin{Bmatrix}{\tilde{p_i}\in\mathbb{R}^{1\times 1\times C}}\end{Bmatrix}^{N_s}_{i=1}$:
\begin{equation}
	\label{eq::prototype}
	\tilde{p_i}=\frac{1}{|G_i|}\sum_{j\in G_i}f_{s,j}, f_{s,j}\in f_s,
\end{equation}
where $G_i$ contains indices of pixels of the $i^{th}$ cluster. In addition, these prototypes are augmented to reflect global context based on an attention mechanism. Formally, part-aware prototypes of a class $c$ from support images are defined as $\mathcal{P}^c_s=\begin{Bmatrix}{p_i}\end{Bmatrix}^{N_s}_{i=1}$:
\begin{equation}
	\label{eq::prototype2}
	p_i=\tilde{p_i}+\lambda_p \sum_{j=1\land j\neq i}^{N_s} a_{i,j}\tilde{p_j}, a_{i,j} = \frac{d(\tilde{p_i},\tilde{p_j})}{\sum_{j\neq i}d(\tilde{p_i},\tilde{p_j})},
\end{equation}
where $\lambda_p$ is a hyperparameter that adjusts the degree of global context reflection and $d(\cdot,\cdot)$ calculates similarity.

After defining prototypes, a query image $I_q$ is segmented based on the similarity between the query features $f_q\in\mathbb{R}^{W'\times H'\times C}$ and $\mathcal{P}^c_s$. Each pixel $f_{q,i,j}\in\mathbb{R}^{1\times 1\times C}$ from $f_q$ calculates cosine-similarity to the prototypes in $\mathcal{P}^c_s$ and selects the nearest one for each class. By aggregating the similarity of each pixel to the nearest, we obtain a similarity map for each class. We then resize this map to the original input size and obtain a softmax probability map. Subsequently, each pixel of $I_q$ is classified to the class of max probability.

\subsection{Neural Uncertainty Estimation}
\label{sec:neuralest}
Our intuition is that uncertainty of prototype-based FSS models mainly stems from various observations of query feature and nearest prototype pairs. To leverage uncertainty in the segmentation task which regresses the probability of pixels belonging to each class, we assume that the probability follows a gaussian distribution similar to prior work~\cite{kendall2017uncertainties,lakshminarayanan2017simple,garnelo2018conditional}. Formally, 
\begin{equation}
	\label{eq::gaussian}
	FSS_{\theta}(\begin{Bmatrix}{I^k_{s}, L^k_{s}}\end{Bmatrix}^{K}_{k=1}, I_{q}) \sim \mathcal{N}(\mu, \sigma^2).
\end{equation}
We estimate $\mu\in\mathbb{R}^{ W\times H\times 2}$ and $\sigma\in\mathbb{R}^{ W\times H\times 2}$ of the gaussian distribution based on the query and nearest prototype feature pairs. Here, $\mu$ is estimated as a similarity-based softmax probability map similar to predictions in PPNet and provides us with a strong baseline. However, as $\mu$ is sometimes overconfident with high probability even though the prototype of a class is not close to the query feature, the model needs a mechanism to capture uncertainty of similarity between the features to produce more reliable predictions. Thus, we propose an uncertainty estimation module $U_w$ (see. Fig.~\ref{fig::train}), that learns a parameter $w$ to estimate $\sigma$ from various observations. This module uses FG/BG prototypes $p^{fg}_s$ and $p^{bg}_s$ and the query feature $f_{q,i,j}$ as inputs when $p^{fg}_s$ and $p^{bg}_s$ are the nearest to $f_{q,i,j}$:
\begin{equation}
	\label{eq::sigma}
	\sigma_{i,j} = U_w(p^{fg}_s, p^{bg}_s, f_{q,i,j}),
\end{equation}
where $i\in [1,W]$ and $j\in [1,H]$. Specifically, the channel sizes of FG/BG prototypes and the query features are first reduced by a linear projection layer, then concatenated and fed into a fully-connected layer block consisting of several linear layers with ReLU activation. Moreover, $\sigma$ is predicted pixel-by-pixel with the final uncertainty map for each class obtained via aggregation of all predictions. The parameters of $E_{\theta}$ and $U_{w}$ are simultaneously optimized to minimize the negative log-likelihood (NLL) loss.
\begin{equation}
	\label{eq::loss}
	\theta, w = \underset{\theta,w}{\operatorname{argmin}}-\sum_{i=1}^W \sum_{j=1}^H \mathcal{N}(L_{q,i,j};\mu_{i,j}, \sigma_{i,j}^2)
\end{equation}
\begin{equation}
	\label{eq::normal_dist}
	\mathcal{N}(L_{q,i,j};\mu_{i,j}, \sigma_{i,j}^2) = \frac{1}{\sigma\sqrt{2\pi}} 
	\exp\left( -\frac{1}{2}\left(\frac{L_{q,i,j}-\mu_{i,j}}{\sigma_{i,j}}\right)^{\!2}\,\right)
\end{equation}

\subsection{Semi-Supervised Few Shot Segmentation}
\label{sec:semisupfss}
In this work, our uncertainty-aware semi-supervised FSS model utilizes pseudo labels of unlabeled images to boost performance. After training $E_{\theta}$ and $U_w$, we define a pseudo label $\hat{L}_u$ of an unlabeled image given the estimates $\mu$ and $\sigma$ and define additional unlabeled data prototypes $\mathcal{P}_u$ from $\hat{L}_u$. The new prototypes provide additional FG/BG information and complement the limited representations of $\mathcal{P}_s$ by capturing varied object part semantics not presented in support images. Even though pseudo-labels are commonly used in semi-supervised approaches, incorrect predictions can deteriorate performance, especially in the FSS task where noisy predictions can lead to using unintended prototypes. To address this, we exclude unreliable predictions from pseudo labels based on uncertainty estimate $\sigma$.

Specifically, given some unlabeled images $I^{1}_{u}$, $I^{2}_{u}$,..., $I^{M}_{u}$ and $\mathcal{P}_s$ obtained from support data, $\mu$ and $\sigma$ of the gaussian distribution are estimated for each $I_u$. 
Though a pseudo label can be simply defined as $\hat{L}_u=round(\mu)$, it may contain incorrect predictions.
Thus, to exclude unreliable predictions from $\hat{L}_u$, we define an uncertainty-aware probability $\mu'$ ranging from $0$ to $1.0$, because both $\mu$ and $\sigma$ have the same range. Herein, 
\begin{equation}
	\label{eq::new_prob}
	\mu' = \mu\times(1-\sigma).
\end{equation}
The obtained probability considers both the initial prediction and uncertainty estimate together. Even though the initial probability of a pixel is high, if its uncertainty is also high, we can obtain $\mu'$ with lower values and vice-versa. Therefore, such pixels will not be included in the uncertainty refined $\hat{L}_u=round(\mu’)$. Consequently, the newly defined pseudo labels only include the pixels with high $\mu$ and low $\sigma$ values. In this way, we effectively reduce the number of incorrect predictions in $\hat{L}_u$. Finally, $\hat{L}_u$ is then used to define prototypes from unlabeled images.

Herein, we proceed to define additional prototypes $\mathcal{P}_u$ using the earlier approach that defines prototypes for support samples. After $\hat{L}_u$ is resized to the same size of the feature map $f_u$, features of FG/BG classes are separated using $\hat{L}_u$ with $N_u$ clusters obtained via K-means clustering. Following, we obtain the mean vector of features belonging to each cluster and consider it as a prototype. For query image segmentation, we use the entire set of prototypes $\mathcal{P}=\mathcal{P}_s \cup \mathcal{P}_u$, and compute the similarity between $f_q$ and each prototype in $\mathcal{P}$ to produce a softmax probability map as segmentation.

\subsection{Implementation Details}
Starting from PPNet as a baseline, we observed that prediction boundaries tend to be inaccurate since it uses the reduced feature map of the last layer of the encoder. To mitigate this, our model additionally trains a refinement module $R$ which refines initial predictions using low-level features similar to CANet~\cite{zhang2019canet}. $R$ intakes three inputs, i.e., global low-level support prototype, low-level query features and initial soft-prediction, which are appropriately resized before concatenation. In particular, $R$ refines the predictions via several convolution layers and a subsequent ASPP module~\cite{chen2017rethinking} without multiple iterations. To effectively use available GPU resources, $R$ was trained separately.

We closely follow the public implementation of PPNet and set the hyperparameters of our model as $\lambda_p=0.8$, and the number of iterations in K-means clustering as $10$. As the authors reported the best performance with 5 clusters in PPNet, we also used 5 clusters in our model, i.e., $N_s=N_u=5$. 

\begin{table}[t]
	\small
	\begin{center}
		\newcolumntype{C}{>{\centering\arraybackslash}m{6em}}
		\begin{tabular}{|l|cccc|}
			\hline
			Methods & \multicolumn{4}{c|}{$M$} \\
			\cline{1-5}
			1-shot & 0 & 3 & 6 &  12 \\
			\hline
			PPNet w $L_{sem}$ & 51.50  & 52.14  & 53.39  & 51.83  \\
			PPNet wo $L_{sem}$ & 51.78  & 52.37  & 53.95  & 52.16  \\
			PPNet + $PL$ & 51.78  & 53.52  & 54.12  & 54.55  \\
			PPNet + $PL$ + $I_{q}$ & - & 53.79  & 54.42  & 54.47  \\
			PPNet + $PL^{\mathcal{H}}$ & 51.78  & 53.26  & 54.00  & 54.50  \\     
			PPNet + $PL^{\mathcal{H}}$ + $I_{q}$ & - & 54.16  & 54.63  & 54.88  \\
			Ours & \bf52.11  & 53.75  & 54.60  & 54.92  \\
			Ours + $I_{q}$ & - & \bf54.88  & \bf55.35  & \bf55.41  \\
			Ours + $I_{q}$ + $R$ & - & 56.27 & 56.70 & 57.13 \\
			\hline
			5-shot &  &  &  &  \\
			\hline
			PPNet w $L_{sem}$ & 62.00  & 62.55  & 63.04  & 62.55  \\
			PPNet wo $L_{sem}$ & 62.02  & 61.40  & 62.14  & 61.36  \\
			PPNet + $PL$ & 62.02  & 61.77  & 62.34  & 62.34  \\
			PPNet + $PL$ + $I_{q}$ & - & 60.82  & 61.89  & 62.06  \\
			PPNet + $PL^{\mathcal{H}}$ & 62.02  & 62.40  & 62.73  & 62.79  \\
			PPNet + $PL^{\mathcal{H}}$ + $I_{q}$ & - & 62.21  & 62.87  & 62.90  \\
			Ours & \bf62.24  & 62.49  & 63.10  & 63.38  \\
			Ours + $I_{q}$ & - & \bf63.20  & \bf63.89  & \bf63.90  \\
			Ours + $I_{q}$ + $R$ & - & 64.73 & 65.27 & 65.38 \\
			\hline
		\end{tabular}
	\end{center}
	\caption{Mean-IoU comparison of the proposed model with different number of unlabeled images against PPNet on PASCAL-$5^i$. RN50 was used as a backbone. $PL$ and $PL^{\mathcal{H}}$ denotes a model using pseudo label and modified pseudo label using $\mathcal{H}$ as uncertainty, respectively. $I_q$ denotes results using query image as an additional $I_u$. Boldface represents the best accuracy without using $R$.}
	\label{test_ppnet}
\end{table}

\begin{table}[t]
	\small
	\begin{center}
		\begin{tabular}{|@{\,}l@{\,}|@{\,}c@{\,}|@{\,}c@{\,}|@{\,}c@{\,}|@{\,}c@{\,}|@{\,}c@{\,}|}
			\hline
			\multirow{2}*{Methods} & \multirow{2}*{$E_{\theta}$} & \multicolumn{2}{c@{\,}|@{\,}}{Mean-IoU} &  \multicolumn{2}{c@{\,}|}{Binary-IoU} \\
			\cline{3-6}
			& & 1-shot & 5-shot & 1-shot & 5-shot \\
			\hline
			CANet~\cite{zhang2019canet} & \multirow{9}*{\shortstack{RN\\50}} & 55.4 & 57.1 & 66.2 & 69.6 \\       
			PGNet~\cite{zhang2019pyramid} &  & 56.0 & 58.5 & 69.9 & 70.5 \\
			PMMs~\cite{yang2020prototype} &  & 55.2 & 56.8 & - & - \\
			PPNet*~\cite{liu2020part} &  & 52.3 & 63.0 & - & - \\
			PFENet~\cite{tian2020prior} &  & 60.8 & 61.9 & \bf73.3 & 73.9 \\
			SAGNN~\cite{xie2021scale} &  & \bf62.1 & 62.8 & 73.2 & 73.3 \\
			ASGNet~\cite{li2021adaptive} &  & 59.3 & 64.4 & 69.2 & 74.2 \\
			Ours &  & 53.6 & 64.3 & 71.0 & 77.3 \\
			Ours* &  & 56.7 & 65.3 & 71.6 & 77.4 \\
			\hline
			FWB~\cite{nguyen2019feature} & \multirow{7}*{\shortstack{RN\\101}} & 56.2 & 59.9 & - & - \\
			DAN~\cite{wang2020few} &  & 58.2 & 60.5 & 71.9 & 72.3 \\
			VPI~\cite{wang2021variational} &  & 57.3 & 60.4 & - & - \\
			PPNet*~\cite{liu2020part} &  & 55.2 & 65.1 & - & - \\
			ASGNet~\cite{li2021adaptive} &  & 59.3 & 64.4 & 71.7 & 75.2 \\
			Ours &  & 55.2 & 65.3 & 72.4 & 78.5 \\
			Ours* &  & 57.4 & \bf67.2 & 73.1 & \bf79.5 \\
			\hline
		\end{tabular}
	\end{center}
	\caption{Comparison of the proposed model against state-of-the-art FSS models on PASCAL-$5^i$. Scores of the comparison methods are taken from literature. ``*" denotes semi-supervised result using 6 unlabeled images.}
	\label{test_pascal}
\end{table}

\section{Experiments}
\subsection{Experimental Setting}
We evaluated the proposed model on commonly used FSS benchmarks, PASCAL-$5^i$~\cite{shaban2017one} and COCO-$20^i$~\cite{nguyen2019feature}. PASCAL-$5^i$ and COCO-$20^i$ have 20 and 80 classes split into 4 folds with 5 and 20 classes each, respectively. We validated our model on the standard 4-fold cross-validation setting. Moreover, every image and its annotation were resized to $(417,417)$ for training and testing. ImageNet~\cite{russakovsky2015imagenet} pre-trained Resnet-50 (RN50) and Resnet-101 (RN101)~\cite{he2016deep} backbones were used for the encoder. 
We follow the evalutation setting in \cite{wang2019panet} which uses mean-IoU and binary-IoU as evaluation metrics. 

We evaluated our model in both supervised and semi-supervised 1-way 1,5-shot settings. In the supervised setting, the model only uses support images to segment a query image without unlabeled images, i.e., the estimated $\mu$ was used as the final predicted probability. In the semi-supervised setting, $6$ unlabeled images were used for comparison against state-of-the-art methods. As our proposed model defines prototypes similar to PPNet, we reproduced PPNet experiments on PASCAL-$5^i$ dataset using public code and considered it as a baseline. Moreover, since one can also use $I_q$ as part of the unlabeled images set, we equally verify whether this setting further boosts performance. 

\begin{figure*} [t]
	\begin{center}
		\includegraphics[width=1.0\linewidth] {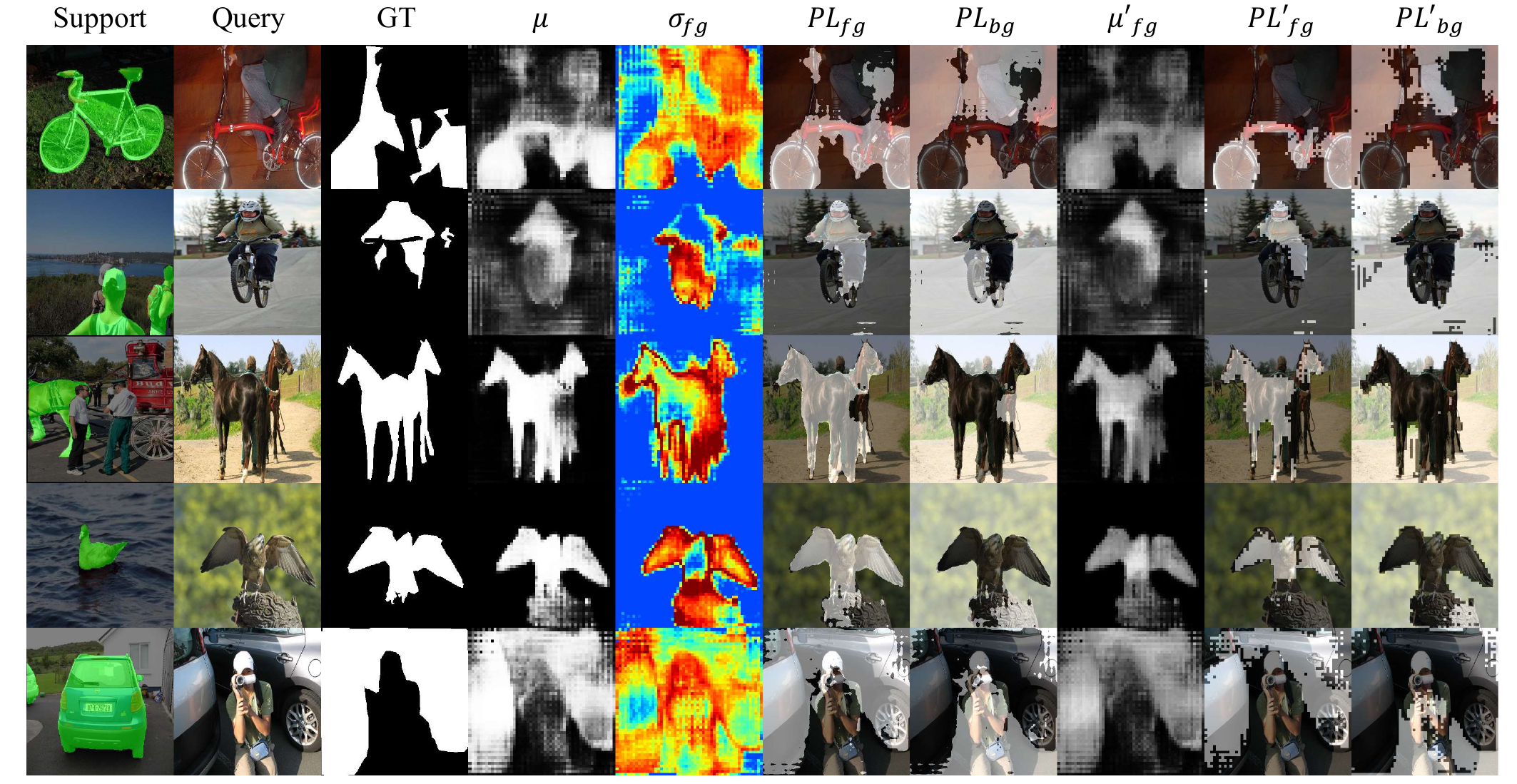}
	\end{center}
	\caption{Visualization of intermediate results in our proposed 1-shot model on PASCAL-$5^i$. Color in $\sigma$ changes from blue to red with higher intensity. $PL_{fg}$ and $PL_{bg}$ denote pseudo labels based on $\mu$ while $PL'_{fg}$ and $PL'_{bg}$ denote pseudo labels based on $\mu'$. }
	\label{fig::qualitative_pascal}
\end{figure*}

\subsection{Results}
\subsubsection{Comparison with PPNet}
In Table~\ref{test_ppnet}, we present the mean-IoU comparison of our method against the reproduced baseline PPNet which is the only semi-supervised FSS model to the best of our knowledge. In the semi-supervised scenario, PPNet was trained once using $M=6$ and tested with different $M$, and later compared to our model without using $R$. Without unlabeled images, our model reports performance on par with PPNet with slight improvements in both 1- and 5-shot settings. As opposed to the cross-entropy loss used in PPNet that forces probabilities to tend to be either 0.0 or 1.0, our formulation better handles ambigious predictions by allowing soft probabilities. 
It is worth noting that semantic regularization $L_{sem}$ proposed by PPNet did not report consistent improvements with different $M$. Thus, we omitted $L_{sem}$ in our framework. Though the best performance reported in the PPNet paper was obtained using $M=6$, improvements were limited as $M$ increases. 

Further, we tested a model that uses additional prototypes based on pseudo labels $\hat{L}_u=round(P)$ where $P$ is the PPNet prediction. In this case, additional unlabeled prototypes $\mathcal{P}_u$ were obtained using the proposed method. We observed that pseudo labels obtained using PPNet to define $\mathcal{P}_u$ could not improve performance as incorrect predictions are included in pseudo labels. We also evaluated whether entropy $\mathcal{H}=-\sum(P\log P)$ is comparable to the proposed uncertainty estimation as in \cite{saporta2020esl}. Herein, the pseudo labels predicted by PPNet were modified using $\mathcal{H}$ instead of $\sigma$, i.e., $\hat{L}_u=round(P\times(1-\mathcal{H}))$. In this case, we noted  marginal improvements over vanilla PPNet $w/ L_{sem}$. This shows that using $\mathcal{H}$ as uncertainty in FSS is helpful but is non-trivial to remove overconfident incorrect predictions in the pseudo labels. Interestingly, all models report better performance when $I_q$ is used together with $I_u$. Overall, our proposed method reports +3.29\% and +1.66\% in 1-shot and 5-shot with $M=12$ using $I_q$, and shows a continual trend as more $I_u$ samples were employed. Further, we obtained higher performances using $R$.

\begin{table}[t]
	\small
	\begin{center}
		\begin{tabular}{|@{\,}l@{\,}|@{\,}c@{\,}|@{\,}c@{\,}|@{\,}c@{\,}|@{\,}c@{\,}|@{\,}c@{\,}|}
			\hline
			\multirow{2}*{Methods} & \multirow{2}*{$E_{\theta}$} & \multicolumn{2}{c@{\,}|@{\,}}{Mean-IoU} &  \multicolumn{2}{c@{\,}|}{Binary-IoU} \\
			\cline{3-6}
			& & 1-shot & 5-shot & 1-shot & 5-shot \\
			\hline
			PPNet*~\cite{liu2020part} & \multirow{4}*{\shortstack{RN\\50}} & 29.0  & 38.5  & - & - \\
			ASGNet~\cite{li2021adaptive} &  & 34.6  & 42.5  & 60.4  & 67.0  \\
			Ours &  & 27.9  & 38.7  & 65.4  & 70.5  \\
			Ours* &  & 31.1  & 41.0  & 66.2  & 70.6  \\
			\hline
			FWB~\cite{nguyen2019feature} & \multirow{7}*{\shortstack{RN\\101}} & 21.2  & 23.7  & - & - \\
			PMMs~\cite{yang2020prototype} &  & 29.6  & 34.3  & - & - \\
			DAN~\cite{wang2020few} &  & 24.4  & 29.6  & 62.3  & 63.9  \\
			PFENet~\cite{tian2020prior} &  & 32.4  & 37.4  & 58.6  & 61.9  \\
			SAGNN~\cite{xie2021scale} &  & \bf37.2  & 42.7  & 60.9  & 63.4  \\
			Ours &  & 31.0  & 40.5  & \bf66.7  & 71.7  \\
			Ours* &  & 32.8  & \bf43.9  & 66.6  & \bf72.6  \\
			\hline
		\end{tabular}
	\end{center}
	\caption{Comparison of the proposed model against state-of-the-art FSS models on COCO-$20^i$. Scores of the comparison methods are taken from literature. ``*" denotes semi-supervised result using 6 unlabeled images.}
	\label{test_coco}
\end{table}
\subsubsection{Comparison with State-of-the-Art Models}
In Table~\ref{test_pascal} and Table~\ref{test_coco}, we report the overall mean-IoU and binary-IoU comparison of our model against other state-of-the-art approaches on PASCAL-$5^i$ and COCO-$20^i$. All reported scores of our model include the refinement module $R$ using $I_q$ as additional $I_u$. On PASCAL-$5^i$, our model with RN50 beats the baseline (PPNet*) even without using unlabeled images, i.e., +1.3\% mean-IoU in both 1-shot and 5-shot settings (Ours). When $6$ unlabeled images were employed, we observed a further boost, i.e., +2.2\% and +1.9\% mean-IoU in 1-shot and 5-shot with RN101, with similar observations on COCO-$20^i$ (Ours*). Interestingly, our method achieved the best scores in the 5-shot setting for both backbones.
Though mean-IoU scores of our 1-shot model was second to that of SAGNN and DAN with different backbones (Table.~\ref{test_pascal}), we report the best 1-shot binary-IoU score on COCO-$20^i$. 
Relatively lower performance of our 1-shot model may be attributed to the weak baseline model. Thus, we believe that the 1-shot model will achieve higher scores if a better baseline is used.

\subsubsection{Qualitative Results}
Fig.~\ref{fig::qualitative_pascal} shows intermediate results of our proposed model. We compare the quality of pseudo labels from $\mu$ and $\mu'$ considering uncertainty. We normalize $\sigma$ between $[0,0.5]$ to be a heatmap since the pixels with $\sigma$ larger than 0.5 are excluded from the pseudo label regardless of $\mu$ using Eq.~(\ref{eq::new_prob}).
Results show that $\sigma$ is high on some ambiguous pixels due to the limited context in the support data. For example, in the first row, the man's leg is falsely classified as FG because its position is near the saddle in the support image. However, in $\mu'$, we were able to suppress such spurious activations for better segmentation in $\hat{L}^{fg}_u$. These results verify that our uncertainty-aware learning model is accurately estimating $\mu$ and $\sigma$.

\begin{figure} [t]
	\begin{center}
		\includegraphics[width=1.0\linewidth] {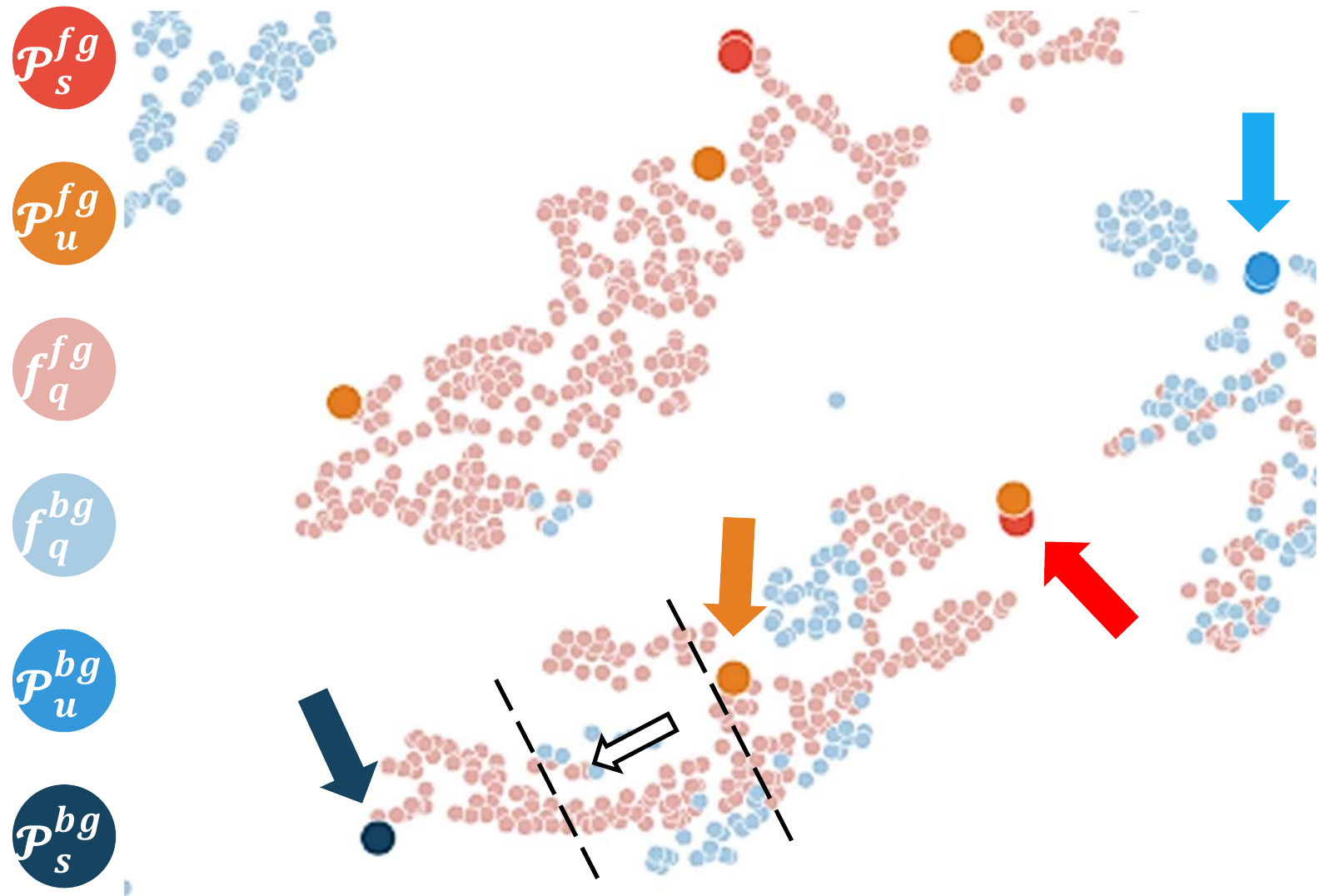}
	\end{center}
	\caption{A t-SNE visualization of query features and prototypes from support and unlabeled images.}
	\label{fig::tsne1}
\end{figure}

Moreover, we show a t-SNE visualization~\cite{van2008visualizing} of query features, and prototypes from support and unlabeled images in Fig.~\ref{fig::tsne1}. 
Here, $f^{fg}_q$ and $f^{bg}_q$ were separated using the true label. As shown in the figure, $\mathcal{P}_u$ provides rich representations relavent to the query features in metric space and supplements the limited context in $\mathcal{P}_s$. In particular, we observed that the decision boundary (dotted line) moves to include more $f^{fg}_q$ by utilizing $\mathcal{P}^{fg}_u$ (orange arrow) while the original decision boundary calculated based on $\mathcal{P}^{fg}_s$ (red arrow) and $\mathcal{P}^{bg}_s$ (navy arrow) causes significant errors. Besides, $\mathcal{P}^{bg}_u$ (blue arrow) provides useful information to classify ambiguous $f^{bg}_q$ which are far from $\mathcal{P}^{bg}_s$. This result shows that newly defined prototypes from unlabeled images are appropriately used for better prediction.

\section{Conclusion}
In this paper, we introduced a novel semi-supervised FSS model which defines additional prototypes from unlabeled images. Our approach also incorporates an uncertainty estimation module tailored for FSS using representations of the query and its nearest prototype pairs. Based on uncertainty estimation, we show that noisy/overconfident pseudo labels obtained from unlabeled data can be refined using estimates  for better FSS performance. Extensive quantitative and qualitative results on popular benchmarks show the effectiveness of our approach over state-of-the-art models. We believe that our semi-supervised learning concept can be generally used in prototype-based FSS models to further improve performance.

\bibliography{aaai22}

\end{document}